\documentclass[twoside,11pt]{article}

\usepackage[T1]{fontenc}
\usepackage[utf8]{inputenc}

\usepackage{jmlr2e}
\usepackage{enumitem}
\usepackage{url}            % simple URL typesetting

\usepackage{booktabs,multirow} % for professional tables
\usepackage{amsmath}
\usepackage{makecell}
\usepackage[export]{adjustbox}
\usepackage{bbm}

\usepackage{pgffor}
\usepackage{amssymb}
\usepackage{comment}
\usepackage{tikz}
\usepackage{asypictureB}
\usepackage{bm}
\usepackage{caption}
\usepackage{wrapfig}
\usetikzlibrary{backgrounds}
\usetikzlibrary{calc}
\usetikzlibrary{fit}
\usetikzlibrary{positioning}
\usepackage{enumitem}
\usepackage[acronym,smallcaps,nowarn]{glossaries}
\usepackage{soul}
\usepackage{hyperref}
\usepackage{graphicx} 
\usepackage{sidecap}

\usepackage[font={footnotesize}]{caption}
\usepackage{todonotes}
\usepackage{algorithm}
\usepackage{float,pbox}

\usepackage{mathtools} %amsthm

\usepackage{multirow}
\usepackage{subfigure}
\usepackage{dcolumn}
\usepackage{algorithmic}
\usepackage{cleveref}
\usepackage{pgfplots}
\usepackage{setspace}
\definecolor{mygreen}{HTML}{167dde}
\definecolor{myred}{HTML}{f22835}
\colorlet{greenfill}{blue!50!white}
\colorlet{purplefill}{red!50!blue!30!white}

\colorlet{redfill}{red!50!white}
\colorlet{moreredfill}{myred!40!white}
\usetikzlibrary{positioning}
\usetikzlibrary{backgrounds}
\usetikzlibrary{calc}
\usetikzlibrary{fit}

\usepackage{pifont}% http://ctan.org/pkg/pifont
\usepackage{pifont}% http://ctan.org/pkg/pifont

\usepackage{tikz}
\usetikzlibrary{fadings}

\usepackage{epsf}

\usepackage{psfrag}
\usepackage{graphicx}
\usepackage{amssymb}
\usepackage{amsbsy}
\usepackage{epsfig}
\usepackage{ifthen}
\usepackage{eucal}
\usepackage{theorem}

\newcommand{\commentout}[1]{}

\newcommand{\B}[1]{\mathbf{#1}}

\newcommand{\PA}{\B{PA}}

  \mathchardef\ordinarycolon\mathcode`\:
  \mathcode`\:=\string"8000
  \begingroup \catcode`\:=\active
    \gdef:{\mathrel{\mathop\ordinarycolon}}
  \endgroup

\usepackage{amsmath,amsfonts,bm}

\def\1{\bm{1}}

\DeclareMathAlphabet{\mathsfit}{\encodingdefault}{\sfdefault}{m}{sl}
\SetMathAlphabet{\mathsfit}{bold}{\encodingdefault}{\sfdefault}{bx}{n}

\usepackage{makecell}
\usepackage{environ}
\usepackage{xcolor}
\usepackage[tikz]{bclogo}
\usepackage{tikz}
\usetikzlibrary{calc}
\usepackage{lipsum}

\NewEnviron{myremark}[1]
  {\par\medskip\noindent
  \begin{tikzpicture}
    \node[inner sep=0pt] (box) {\parbox[t]{.99\textwidth}{%
      \begin{minipage}{.20\textwidth}
      \centering\tikz[scale=1]\node[scale=2,rotate=0]{\bclampe};
      \end{minipage}%
      \begin{minipage}{.75\textwidth}
      \textbf{#1}\par\smallskip
      \BODY
      \end{minipage}\hfill}%
    };
    \draw[red!75!black,line width=3pt] 
      ( $ (box.north east) + (-5pt,3pt) $ ) -- ( $ (box.north east) + (0,3pt) $ ) -- ( $ (box.south east) + (0,-3pt) $ ) -- + (-5pt,0);
    \draw[red!75!black,line width=3pt] 
      ( $ (box.north west) + (5pt,3pt) $ ) -- ( $ (box.north west) + (0,3pt) $ ) -- ( $ (box.south west) + (0,-3pt) $ ) -- + (5pt,0);
  \end{tikzpicture}\par\medskip%
}

\ShortHeadings{Higher-Level Cognitive Inductive Biases}{Higher-Level Cognitive Inductive Biases}
\firstpageno{1}

\begin{document}

%AG: Which one is better ?
\title{Inductive Biases for Deep Learning of Higher-Level Cognition}
%\title{Inductive Biases for Deep Learning of Conscious Processing}
% Moving from Deep Statistical Models to Deep Structural Mechanistic Models by using Higher-Level Cognitive Inductive Biases}

%Motivating the rules of the game for deep structural mechanisms models using insights from higher level cognitive inductive biases 

%\title{Structured Deep Learning: Attention, Sparsity, Modularity, and Lessons from cognitive science}
% \title{Language of Generalization}

%Alex outline
% 

\author{\name Anirudh Goyal \email anirudhgoyal9119@gmail.com \\
       \addr Mila, University of Montreal\\
       \AND
       \name Yoshua Bengio \email yoshua.bengio@mila.quebec \\
       \addr Mila, University of Montreal}

%\author{\name Anirudh Goyal, \name Yoshua Bengio}

\editor{}

\maketitle

\begin{abstract}%   <- trailing '%' for backward compatibility of .sty 

A fascinating hypothesis is that human and animal intelligence could
be explained by a few principles (rather than an encyclopedic list of heuristics). If that hypothesis was correct, we could more easily both understand our own intelligence and build intelligent machines. Just like in physics, the principles themselves
would not be sufficient to predict the behavior of complex systems
like brains, and substantial computation might be needed to simulate
human-like intelligence. 
This hypothesis would suggest
that studying the kind of inductive biases
that humans and animals exploit could help both clarify these principles
and provide inspiration for AI research and neuroscience theories. Deep learning
already exploits several key inductive biases, and this work considers
a larger list, focusing on those which concern mostly higher-level
and sequential conscious processing. The objective of clarifying these particular 
principles is that they could potentially help us build AI systems 
benefiting from humans' abilities in terms of flexible out-of-distribution and systematic generalization, which
is currently an area where a large gap exists between
state-of-the-art machine learning and human intelligence.

\end{abstract}

\section{Has Deep Learning Converged?}

Is 100\% accuracy on the test set enough?  Many machine learning systems have achieved excellent accuracy across a variety of tasks \citep{deng2009imagenet, mnih2013playing, schrittwieser2019mastering}, yet the question of whether their reasoning or judgement is correct has come under question, and answers seem to be wildly inconsistent, depending on the task, architecture, training data, and interestingly, the extent to which test conditions match the training distribution.  Have the 
main principles required for deep learning to achieve human-level performance
been discovered, with the main remaining obstacle being to scale up?
Or do we need to follow a completely different research direction not built on the principles discovered with deep learning, in order to achieve the kind of cognitive competence displayed  by humans? Our goal here is to better understand the gap between current deep learning and human cognitive abilities so as to help answer these questions and suggest research directions for deep learning with the aim of bridging the gap towards human-level AI. Our main hypothesis is that deep learning succeeded in part because of a set of inductive biases (preferences, priors or assumptions), but that additional ones should be included in order to go from good in-distribution  generalization in highly supervised learning tasks (or where strong and dense rewards are available), such as object recognition in images, to strong out-of-distribution generalization and transfer learning to new tasks with low sample complexity (few examples needed to generalize well). To make that concrete, we consider some of the inductive biases humans may exploit in conscious thought using highly sequential cognition operating at the level of conscious processing, and review some early work exploring these ``high-level cognitive inductive priors'' in deep learning. We use the term {\em high-level} to talk about variables that are manipulated at the conscious level of processing and are thus generally verbalizable. However, humans can consciously focus attention on low-level or intermediate-level features, e.g., by describing an odd-coloured pixel, not just very abstract concepts like objects or social situations. We argue that the deep learning progression from MLPs to convnets  to transformers has in many ways been an (incomplete) progression towards the original goals of deep learning, i.e., to enable the discovery of a hierarchy of representations, with the most abstract ones, often associated with language, at the top. Note however, that although language may give us a view on system 2, these abilities are likely to pre-exist language as there is evidence of surprisingly strong forms of on-the-fly reasoning in some non-human animals, like corvids~\citep{taylor2009new}. Our arguments suggest that while deep learning brought remarkable progress, it  needs to be extended in qualitative and not just quantitative ways: larger and more diverse datasets and more computing resources \citep{brown2020language} are important but insufficient without additional inductive biases \citep{vaswani2017attention, he2016deep, gilmer2017neural, shazeer2017outrageously,  fedus2021switch, hinton2021represent, welling2019we, dosovitskiy2020image, battaglia2018relational}.  We make the case that evolutionary forces, the interactions between multiple agents, the non-stationary and competition systems put pressure on the learning machinery to achieve the kind of flexibility, robustness and ability to adapt quickly which humans seem to have when they are faced with new environments \citep{bansal2017emergent, liu2019emergent, baker2019emergent, leibo2019autocurricula} but needs to be improved with deep learning. The sought-after inductive biases should thus especially help AI to progress on these fronts. In addition to thinking about the learning and sample complexity advantage of these inductive biases, this paper links them with knowledge representation in neural networks, with the idea that by decomposing knowledge into its stable parts (like causal mechanisms) and volatile parts (random variables), and factorizing knowledge in small and somewhat independent pieces that can be recomposed dynamically as needed (to reason, imagine or explain at an \textit{explicit} and verbalizable level), one may achieve the kind of systematic generalization which humans enjoy and is common in natural language~\citep{marcus1998rethinking, marcus2019algebraic, lake2017generalization, bahdanau2018systematic, mcclelland1987parallel}.

%Progress in Deep Learning has largely fulfilled the initial connectionist vision of learning to connect and weight neurons based on feedback from data.  The system which most embodies this ideal is the simple fully-connected MLP, in which all neurons are wired together within an example, and all connections are learned directly from data.  At the same time, the most successful deep learning systems have introduced additional structure.  For example, convolutional neural networks share parameters across space, and only have local connections.  The increasingly popular attention mechanisms (including transformers) learn to dynamically share information between different positions.  Thus it is clear that while flexibility and learning are important, using deep networks which have a structured connectivity is critical for better performance. \Ani{better performance with respect to what ?}  

\vspace{-3mm}
\subsection{Data, Statistical Models and Causality}
\label{sec:data}

Our current state-of-the-art machine learning systems sometimes achieve good performance on a specific and narrow task, using very large quantities of labeled data, either by supervised learning or reinforcement learning (RL) with strong and frequent rewards. Instead, humans are able to understand their environment in a more unified way (rather than with a separate set of parameters for each task) which allows them to quickly generalize (from few examples) on a new task, thanks to their ability to reuse previously acquired knowledge. Instead, current systems are generally not robust to changes in distribution~\citep{peters2017elements, geirhos2020shortcut, hendrycks2021many, koh2021wilds, schneider2020improving}, adversarial examples~\citep{goodfellow2014explaining, kurakin2016adversarial}, spurious correlations \citep{krueger2021out, beery2018recognition, arjovsky2019invariant} etc.   

One possibility studied in the machine learning literature is that we should train our models with multiple datasets, each providing a different view of the underlying model of the world shared by humans \citep{baxter2000model}. Whereas multi-task learning usually just pools the different datasets \citep{caruana1997multitask, collobert2008unified,  ruder2017overview}, we believe that there is something more to consider: we want our learner to perform
well on a completely new task or distribution, either immediately (with zero-shot out-of-distribution generalization), or with a few examples (i.e. with efficient transfer learning) \citep{ravi2016optimization, wang2016learning, finn2017model, cabi2019scaling,  jang2022bc, reed2022generalist, ahn2022can, brown2020language, alayrac2022flamingo, borgeaud2022improving, chowdhery2022palm, sanh2021multitask, lu2022unified, raffel2020exploring}.

This raises the question of changes in distribution or task. Whereas the traditional train-test scenario and learning theory assumes that test examples come from the same distribution as the training data, just dropping that assumption means that we cannot say anything about generalization to a modified distribution. Hence new assumptions are required about how the different tasks or the different distributions encountered by a learning agent are related to each other.

We use the term {\em structural-mechanistic}~\citep{Schoelkopf15} to characterize models which follow an underlying mechanistic understanding of reality.  They are closely related to the
structural causal models used to capture causal structure~\citep{Pearl2009}.
The key property of such models is that they will make correct predictions over a variety of data distributions which are drawn from the same underlying causal system, rather than being specific to a particular distribution.  To give a concrete example, the equation $E = MC^2$ relates mass and energy in a way which we expect to hold regardless of other properties in the world.  On the other hand, an equation like "${\rm GDP}_{t} = 1.05\, {\rm GDP}_{t-1}+\,$noise" may be correct under a particular data distribution (for example a country with some growth pattern) but will fail to hold when some aspects of the world are changed, even in ways which did not happen or could not happen, i.e., in a counterfactual.

However, humans do not represent all of their knowledge in such a neat verbalizable way as Newton's equations. Most humans understand physics first of all at an \textit{intuitive} level and in solving practical problems we typically combine such implicit knowledge with explicit verbalizable knowledge \citep{mccloskey1983intuitive, baillargeon1985object, spelke1992origins, battaglia2013simulation}. We can name high-level variables like position and velocity but may find it difficult to explain the intuitively known mechanisms  which relate them to each other, in everyday life (by opposition to a physicist running a simulation of Newton's equations). 

%An important question for us is how knowledge can be represented in these two forms, the implicit --intuitive and difficult to verbalize -- and the explicit -- which allows humans to share part of their thinking process through natural language. 

\begin{myremark}{Implicit and Explicit Knowledge}
An important question for us is how knowledge can be represented in these two forms, the implicit --
intuitive and difficult to verbalize -- and the explicit -- which allows humans to share part of their thinking process through natural language. 
\end{myremark}

Causal understanding hinges on capturing the effect of interventions as changes in distribution.
Humans frequently explain their perception (at the explicit level) and reason in terms of causal structure, and causal structure is really about how a joint distribution between causal random variables can change under interventions, i.e., actions.
This suggests that one possible direction that deep learning needs to incorporate includes more notions about agency, reasoning and causality, even when the application only involves single inputs like an image and not actually learning a policy. For this purpose we need to examine how to go beyond the statistical learning framework that has dominated deep learning and machine learning in recent decades. Instead of thinking of data as a set of examples drawn independently from the same distribution, we should probably reflect on the origin of the data through  a real-world non-stationary process. We claim that this perspective would help learning agents, such as babies or robots to succeed in the changing environments.  This paper mostly discusses inductive biases inspired by higher-level cognition and aimed at facing these generalization challenges, pointing to existing work to implement some of them. However, for the most part, how to efficiently implement and combine these inductive biases in a single system remains an open question.

\vspace{-3mm}
\section{About Inductive Biases}
\label{sec:about-inductive-biases}

The no-free-lunch theorem for machine learning~\citep{wolpert1995no, baxter2000model} basically says that some set of preferences (or inductive bias) over the space of all functions is necessary to obtain generalization, that there is  no completely general-purpose learning algorithm, that any learning algorithm  will generalize better on some distributions and worse on others. Typically, given a particular dataset and loss function, there are many possible solutions (e.g. parameter assignments) to the learning problem that exhibit equally ``good'' performance on the training points. Given a finite training set, the only way to generalize to new input configurations is then to rely on some assumptions or preferences about the solution we are looking for. An important question for AI research aiming at human-level performance then is to identify inductive biases that are most relevant to the human perspective on the world around us. Inductive biases, broadly speaking, encourage the learning algorithm to prioritise solutions with certain properties.
Table~\ref{tab:current-inductive-biases} lists some of the inductive biases already used in various neural networks, and the corresponding properties. Although they are often expressed in terms of a neural architecture, they can also be about how the networks are trained, e.g., unsupervised pre-training, self-supervised learning and semi-supervised training, which
all have to do with the input distribution $P(X)$ being informative about future tasks $P(Y|X)$. Other relevant elements
which are not directly about inductive biases (and not discussed further in this paper) include for example the ability of a learning agent to actively seek knowledge (e.g. in active learning or reinforcement learning)  or to obtain information from other agents (e.g., social learning, multi-agent learning).

\begin{table}[]
    \centering
    \begin{tabular}{|c|c|}
      \hline
       \textbf{Inductive Bias} & \textbf{Corresponding property}  \\
       \hline
       Distributed representations & Inputs mapped to patterns of features \\
       \hline
       Convolution & group equivariance (usually over space) \\ 
       \hline
       Deep architectures & Complicated functions = composition of simpler ones \\
       \hline
       Graph Neural Networks &  equivariance over entities and relations \\
       \hline
       Recurrent Nets & equivariance over time  \\
        \hline 
        Soft attention & equivariance over permutations  \\
        \hline Self-supervised pre-training & $P(X)$ is informative about $P(Y|X)$ \\ \hline
    
    \end{tabular}
    \caption{Examples of current inductive biases in deep learning. Many 
    have to do with the architecture while the last one influences
    the training framework and objective.}
    \label{tab:current-inductive-biases}
\end{table}

%Efficient inference can be achieved by sequentially considering one or few factors at a time (with attention

\begin{table}[]
    \centering
    \scalebox{0.7}{\begin{tabular}{|c|c|c|}
      \hline
       \textbf{Inductive Bias}  & \textbf{Corresponding Property} & \textbf{Relevant References}  \\
       \hline
      \makecell{High-level variables\\ play a {\bf causal} role} & \makecell{Learning representations of \\ latent entities/attributes} & \makecell{\citep{le2011learning, eslami2016attend, greff2016tagger}\\ \citep{raposo2017discovering, engelcke2019genesis} \\ \citep{kosiorek2018sequential, goyal2020object, ke2021systematic}\\ \citep{greff2019multi, locatello2020object, burgess2019monet} \\ \citep{ahmed2020causalworld, goyal2019recurrent, zablotskaia2020unsupervised}\\  \citep{rahaman2020s2rms, du2020unsupervised, ding2020object} \\ \citep{goyal2021coordination, van2018relational}}. \\
      \hline
      \makecell{Changes in distribution are \\ due to causal {\bf interventions} } & \makecell{The source of changes in distribution \\ is sparse and localized in the \\ appropriate semantic space} & \makecell{\citep{bengio2019meta, scholkopf2012causal, scholkopf2021toward} \\ \citep{ke2020amortized, goyal2019recurrent, ke2019learning}} \\
      \hline
      \makecell{{\bf Knowledge is generic}, defined \\ over abstract variables, and can \\ be applied on different \\ instances} & \makecell{Factorizing knowledge in terms of \\ abstract variables and reusable functions\\ that encapsulate how these variables\\ interact with each other} & \citep{alias2021neural, ke2021systematic, silver2022inventing} \\
      \hline
      \makecell{{\bf Sparsity of dependencies}}  & \makecell{Learned functions operate on \\ a sparse set of variables \\ (like arguments in \\ typed-programming languages)} & \citep{alias2021neural, ke2021systematic} \\
       \hline
      \makecell{{\bf Short causal chains}} & \makecell{Causal chains used to perform learning \\ or inference (to obtain explanations \\ for achieving some goal) are broken\\ down into short causal chains of events\\ that may be far in time from each other} &\makecell{\cite{ke2018sparse, lillicrap2020backpropagation, arjona2019rudder} \\ \citep{patil2020align, kerg2020untangling, goyal2022retrieval}} \\
      \hline
      \makecell{Context-dependent processing \\{\bf involving goals, top-down} \\ {\bf influence, and bottom-up} \\ competition}  & \makecell{Top-down contextual information
is \\ dynamically combined with bottom-up \\ sensory signals at every level \\
of the hierarchy of computations \\ relating low-level and \\ high-level representations} & \makecell{\cite{mittal2020learning, fan2020addressing} \\ \cite{mcclelland2020placing}} \\
     \hline
    \end{tabular}}
    \caption{Proposed additional inductive biases for deep learning: much progress has been made in learning representation of high level variables (entities or objects). Much more progress is needed on other inductive biases such as the ones listed above. It would also be useful to integrate these inductive biases into a unified architecture. }
    \label{tab:proposed-inductive-biases}
\end{table}

\vspace{-2mm}
\paragraph{From Inductive Biases to Algorithms.}
There are many ways to encode such biases—e.g. explicit regularisation objectives \citep{bishop1995neural, bishop1995training, srivastava2014dropout,  kukavcka2017regularization, zhang2017mixup}, architectural constraints \citep{yu2015multi,long2015fully, dumoulin2016guide, he2016deep,  huang2017densely}, parameter sharing \citep{hochreiter1997long, pham2018efficient},
implicit effects of the choice of the optimization method \citep{jastrzkebski2017three, smith2017bayesian, chaudhari2018stochastic}, self-supervised learning or self-supervised pre-training \citep{hinton2006fast, erhan2010does, devlin2018bert, chen2020simple, chen2020improved, grill2020bootstrap}, invariance or equivariance to known transformations \citep{bruna2013spectral, defferrard2016convolutional, ravanbakhsh2017equivariance, thomas2018tensor, finzi2020generalizing,  satorras2021n} or choices of prior distributions in a Bayesian model \citep{jeffreys1946invariant, berger1992development, gelman1996bayesian, fortuin2022priors}. For example, one can build translation invariance of a neural network output by replacing matrix multiplication by convolutions \citep{lecun1995convolutional} and pooling \citep{krizhevsky2012imagenet}, or by averaging the network predictions over transformations of the input (feature averaging) \citep{zhang2017mixup}, or by training on a dataset augmented with these transformations (data augmentation) \citep{krizhevsky2012imagenet}.  Whereas some inductive biases can easily be encoded into the learning algorithm (e.g. with convolutions), the preference over functions is sometimes implicit and not intended by the designer of the learning system, and it is sometimes not obvious how to turn an inductive bias into a machine learning method, this conversion often being the core contribution of machine learning papers.

\vspace{-2mm}
\paragraph{Inductive Biases as Data.}
We can think of inductive biases or priors and built-in structure as ``training data in disguise'', and one can compensate lack of sufficiently powerful priors by more data \citep{welling2019we}. Interestingly,  different inductive biases may be equivalent to more or less data (even possibly exponentially more data): we suspect that inductive biases based on a form of compositionality (like distributed representations \citep{pascanu2013number}, depth~\citep{montufar2014number} and attention \citep{attention, vaswani2017attention}) can potentially also provide a larger advantage (to the extent that they apply well to the function to be learned). In general, priors can be imperfect and this shows most with large datasets. Even for good priors, the advantage of inductive biases may be smaller on very large datasets, which suggests that transfer settings (where only few examples are available for the new distribution) are interesting to evaluate the advantage of inductive biases and of their implementation.

%\vspace{-2mm}
%\paragraph{Inductive Biases and Ease of Learning.}
%Interestingly, we sometimes observe a link between the advantage that an inductive bias brings in terms of generalization and the advantage it brings in terms of ease of optimization (e.g. attention, or residual connections). This makes sense if one thinks of the training process as a form of generalization inside the training set. With a strong inductive bias, the first half of the training set already strongly constrains what the learner should predict on the second half of the training set, which means that when we encounter these examples, their error is already low and training thus converges faster.

\vspace{-2mm}
\paragraph{Agency, Sequential Decision Making and Non-Stationary Data Streams.}
The classical framework for machine learning is based on the assumption of identically and independently distributed data (i.i.d.), i.e test data has the same distribution as the training data. This is a very important assumption, because if we did not have that assumption, then we would not be able to say anything about generalization to new examples from the same distribution. Unfortunately, this assumption is too strong, and reality is not like this, especially for agents taking decisions one at a time in an environment from which they also get observations. The distribution of observations seen by an agent may change for many reasons: the agent acts (intervenes) in the environment, other agents intervene in the environment, or simply our agent is learning and exploring, visiting different parts of the state-space as it does so, discovering new parts of it along the way, thus experiencing non-stationarities along the way. Although sequential decision-making is ubiquitous in real life, there are scenarios where thinking about these non-stationarities
may seem unnecessary (like object recognition in static images). However, if we want to build learning systems which are robust to changes in distribution, it may be necessary to train them in settings where the distribution changes! And then of course there are applications of machine learning where the data is sequential and non-stationary (like historical records of anything) or even more so, where the learner is also an agent or is an agent interacting with other agents (like in robotics, autonomous driving or dialogue systems). That means we may need to go away from large curated datasets typical of supervised learning frameworks and instead construct non-stationary controllable environments as the training grounds and benchmarks for our learners. This complicates the task of evaluating and comparing learning algorithms but is necessary and we believe, feasible, e.g. see~\citep{yu2017preparing,packer2018assessing,chevalier2018babyai,dulac2020empirical,ahmed2020causalworld}.

\vspace{-2mm}
\paragraph{Transfer Learning and Continual Learning. }
Instead of a fixed data distribution and searching for an inductive bias which works well with this distribution, we are thus interested in transfer learning~\citep{pratt1991direct, pratt1993discriminability} and continual learning~\citep{ring1998child} scenarios,  with a potentially infinite stream of tasks,  and where the learner must extract information from  past experiences and tasks to improve its learning speed (i.e., sample complexity, which is different from asymptotic performance which is currently the standard) on future and yet unseen tasks. Suppose the learner faces a sequence of tasks, A, B, C  and then we want the learner to perform well on a new task D. Short of any assumptions it is nearly impossible to expect the learner to perform well on D. However if there is some shared structure, between the transfer task (i.e task D) and source tasks (i.e tasks A, B and C), then it is possible to generalize or transfer knowledge from the source task to the target task.
Hence, if we want to talk meaningfully about knowledge transfer, it is important to talk about the assumptions on the kind of data distribution that the learner is going to face, i.e., (a) what they may have in common, what is stable and stationary across the environments experienced and (b) how they differ or how changes occur from one to the next in case we consider a sequential decision-making scenario. This division should be reminiscent of the work on \textit{ meta-learning}~\citep{bengio1990learning, schmidhuber1987evolutionary, finn2017model, ravi2016optimization}, which we can understand as dividing learning into slow learning
(of stable and stationary aspects of the world) and fast learning (of task-specific aspects of the world).
This involves two time scales of learning, with an outer loop for meta-learning of meta-parameters
and an inner loop for regular learning of regular parameters. In fact we could have more than two time scales \citep{clune2019ai}: think about the outer loop of evolution, the slightly faster loop of cultural learning \citep{bengio2014evolving} which is somewhat stable across generations, the faster learning of individual humans, the even faster learning of specific tasks and new environments within a lifetime, and the even faster inner loops of motor control and planning which adapt  policies to the specifics of an immediate objective like reaching for a fruit. Ideally, we want to build an understanding of the world which shifts as much of the learning to the slower and more stable parts so that the inner learning loops can succeed faster, requiring less data for adaptation.

\vspace{-3mm}
\paragraph{Systematic Generalization and Out-of-Distribution Generalization.} In this paper, we focus on the objective of out-of-distribution (OOD) generalization, i.e., generalizing outside
of the specific distribution(s) from which training observations were drawn. A more general way to conceive of
OOD generalization is with the concept of sample complexity in the face of new tasks or changed distributions.
One extreme is zero-shot OOD generalization while the more general case, often studied in meta-learning setups,
involves k-shot generalization (from $k$ examples of the new distribution).

Whereas the notions of OOD generalization and OOD sample complexity tell us what we want to achieve (and hint at how we might measure it) they say nothing about how to achieve it. This is where the notion of \textit{systematic generalization} becomes interesting \citep{smolensky1988proper, fodor1988connectionism, marcus1998rethinking, mcclelland1987parallel}. Systematic generalization is a phenomenon which was
first studied in linguistics~\citep{lake2017generalization, bahdanau2018systematic} because it is a core property
of language: the meaning for a novel composition of existing concepts (e.g. words) can be derived
systematically from the meaning of the composed concepts. This very clearly exists in language, but 
humans benefit from it in other settings, e.g., understanding a new object by combining properties of different parts which compose it. Systematic generalization even makes it possible to generalize to new combinations that have zero probability under the training distribution: it is not just that they did not occur in the training data, but that even if we had seen an infinite amount of training data from our training distribution, we would not have any sample showing this particular combination. For example, when you read a science fiction scenario for the first time, that scenario could be impossible in your life, or even in the aggregate experiences of billions of humans living today, but you can still imagine it and make sense of it (e.g., predict the end of the scenario from the beginning). Empirical studies of systematic generalization were performed by~\citep{bahdanau2018systematic, bahdanau2019closure}, where particular forms of combinations of linguistic concepts were present in the training distribution but not in the test distribution, and current methods take a hit in performance, whereas humans would be able to answer such questions easily. 
 
Humans use inductive biases providing forms of compositionality, making it possible to  generalize from a finite set of combinations to a larger set of combinations of concepts. Deep learning already benefits from a form of compositional advantage with distributed representations~\citep{hinton1984distributed,bengio2000taking,bengio2001neural}, which are at the heart of why neural networks work so well. There are theoretical arguments about why distributed representations can bring a potentially exponential  advantage~\citep{pascanu2013number}, if this matches properties of the underlying data distribution. Another advantageous form of compositionality in deep nets arises from the depth itself, i.e., the composition of functions, with provable up to exponential advantages under the appropriate assumptions~\citep{montufar2014number}. However, a form of compositionality that we propose here and should
be better incorporated in deep learning is the form called systematicity~\citep{lake2018generalization} defined by linguists, and more recently
systematic generalization in machine learning papers~\citep{bahdanau2018systematic, ruis2020benchmark, akyurek2020learning}.

Current deep learning methods tend to overfit the training {\em distribution}. This would not be visible by looking at a test set from the same distribution as the training set, so we need to change our ways of evaluating the success of learning because we would like our learning agents to generalize in a systematic way, out-of-distribution.  This only makes sense if the new environment has enough shared components or structure with previously seen environments, which corresponds to certain assumptions on distributional changes, bringing back the need for appropriate inductive biases, about distributions (e.g., shared components) as well as about how they change (e.g., via agents' interventions).

The structure of next two sections is as follows: In section ~\ref{sec:inductive_biases}, we motivate some of the system-2 inductive biases inspired by human cognition. We think endowing machine learning systems with an efficient implementation of these inductive biases could improve (there's an ample evidence already but much more progress needs to be made to achieve human-level AI) the generalization and adaptation performance of the machine learning models. In section~\ref{sec:scm}, we  open a parenthesis to review material on causal dependencies to deepen some of the discussion made in section~\ref{sec:inductive_biases} on inductive biases linked to the causal nature of high-level semantic variables. 

\vspace{-2mm}
\section{Inductive biases based on higher-level cognition as a path towards systems that generalize better OOD}
\label{sec:inductive_biases}
 
\paragraph{Synergy between AI research and cognitive neuroscience.}
Our aim is to take inspiration from (and further develop) research into the cognitive science of conscious processing, to deliver greatly enhanced AI, with abilities observed in humans thanks to high-level reasoning leading among other things to greater abilities to face unusual or novel situations by reasoning, compositionally reusing existing knowledge and being able to communicate about that. At the same time, new AI models could drive new insights into the neural mechanisms underlying conscious processing, instantiating a virtuous circle. Machine learning procedures have the advantage that they can be tested
for their effective learning abilities, and in our case in terms of out-of-distribution abilities or in the context
of causal environments changing due to interventions, e.g., as in~\citep{ahmed2020causalworld}. Because they have to be very formal, AI models can also suggest hypotheses
for how brains might implement an equivalent strategy with biological machinery. Testing these hypotheses could in turn
provide more understanding about how brains solve the same problems and help to refine the deep learning systems.

\vspace{-2mm}
\subsection{Conscious vs Unconscious Processing in Brains}
\label{sec:conscious-vs-unconscious}
\label{sec:gwt}

Imagine that you are driving a car from your office, back home. You  do not need to pay a lot of attention to the road and you can talk to the passenger. Now imagine encountering a road block due to construction: you have to pay more attention, you have to be on lookout for new information, if the passengers starts talking to you, then you may have to tell the person, ``please let me drive''. It is interesting to consider that  when humans are confronted with a new situation, very commonly they require their \textit{conscious} attention \citep{carlson1985conscious, newman1997neural}. In the driving example, when there is a road block you need to pay attention in order to think through what to do next, and you probably don't want to be disturbed, because your conscious attention can only focus on one thing at a time. 

There is something in the way humans process information which seems to be different -- both functionally and in terms of the neural signature in the brain -- when we deal with conscious processing and novel situations (changes in the distribution) which require our conscious attention, compared to our habitual routines. In those novel situations, we generally have to \textit{think}, \textit{focus} and \textit{attend} to specific elements of our perception, actions or memories and sometimes inhibit our reactions based on context (e.g., facing new traffic rules or a road block). Why would humans have evolved to deal with such an ability with changes in distribution?  Maybe simply because life experience is highly non-stationary.
%, and the contextual information we exchange with language could drastically change how we should interpret things and act, often having to combine old pieces of knowledge in completely novel ways. \newline

\paragraph{System 1 and System 2.} Cognitive scientists distinguish~\citep{schneider1982automatic, redgrave2010goal, botvinick2001conflict, botvinick2001evaluating, mozer2001rational, bargh1984automatic} \textit{habitual} versus \textit{controlled} processing, where the former correspond to  default behaviors, whereas the latter require attention and \textit{mental effort}. Daniel Kahneman introduced the framework of fast and slow thinking~\citep{kahneman2011thinking}, and describes the \textit{system 1} and \textit{system 2} styles of processing in our brain. Some tasks can be achieved using only system 1 abilities whereas others also require system 2 and conscious processing. There are also notions of explicit (verbalizable) knowledge and explicit processing (which roughly correspond to system 2) and implicit (intuitive) knowledge and corresponding system 1 neural computations. The default (or unconscious) processing of system 1 can take place very rapidly (as fast as about 100ms) and mobilize many areas of the brain in parallel. On the other hand, controlled (or conscious) processing involves a sequence of thoughts, usually verbalizable, typically requiring seconds to achieve. Whereas we can act in fast and precise habitual ways without having to think consciously, the reverse is not true: controlled processing (i.e., system 2 cognition) generally requires unconscious processing to perform much of its work. It is as if the conscious part of the computation was just the top-level program and the tip of the iceberg. Yet, it seems to be a very powerful one, which makes it possible for us to solve new problems creatively by recombining old pieces of knowledge, to reason, to imagine explanations and future outcomes, to plan and to apply or discover causal dependencies. It is also at that level that we interface with other humans through natural language. And when a word refers to a complex concept for which we do not have a clear verbalizable and precise explanation (like how we manage to drive our bike), we can still name it and reason about how it relates with other pieces of knowledge, etc. Even imagination and planning (which are hallmarks of system 2 abilities) require system 1 computations to sample candidate solutions to a problem (from a possibly astronomical number, which we never have to explicitly examine). 

Our brain seems to thus harbour two very different types of knowledge: the kind we can explicitly reason about and communicate verbally (system 2 knowledge) and the kind that is intuitive and implicit (system 1 knowledge). When we learn something new, it typically starts
being represented explicitly, and then as we practice it more, it may migrate to a different, implicit form. When you learn the grammar of a new language, you may be given some set of rules, which you try to apply on the fly, but that requires a lot of effort and is done painfully slowly. As you practice this skill, it can gradually migrate to a habitual form, you make less mistakes (for the common cases), you can read / translate / write more fluently, and you may even eventually forget the original rules. When a new rule is introduced, you may have to move back some of that processing to system 2 computation to avoid inconsistencies. It looks as if one of the key roles of conscious processing is to integrate different sources of knowledge (from perception and memory) in a coherent way. 

\paragraph{The Global Worskpace Theory.}
The above division of labour is at the heart of the cognitive neuroscience Global Workspace Theory (or GWT) from Baars~\citep{baars1993cognitive,baars1997theatre} and its extension, the Global Neuronal Workspace model~\citep{shanahan2006cognitive, shanahan2010embodiment, shanahan2012brain, dehaene2011experimental, Dehaene-et-al-2017,dehaene2020we}. 
The GWT suggests an architecture allowing specialist components to interact.  The key claim of the GWT is the existence of a shared representation---sometimes called a blackboard ~\citep{mcclelland1986programmable}, sometimes a workspace---that can be modified by any selected
specialist and whose content is broadcast to all specialists. That selection is based on a form of attention and can correspond to dynamically  selecting  (based on the input) a module or a few modules in a modular neural net that are most appropriate for a particular context and task.
The basic idea of deep learning frameworks inspired by the GWT is to  explore a similar communication and coordination scheme for a neural net comprising of distinct modules~\citep{shanahan2006cognitive, shanahan2005consciousness}. The GWT theory posits that conscious
processing revolves around a communication bottleneck between selected parts of the brain which are called
upon when addressing a current task. There is a threshold of relevance beyond which information which was previously
handled unconsciously gains access to this bottleneck, instantiated in a working memory. When that happens,
that information is broadcast to the whole brain, allowing the different relevant parts of it to synchronize,
forcing each module to learn to exchange with other modules in a way that allows swapping one module for another as source or destination of communicated content, i.e., with a shared "language". These shared representations can be interpreted by many other modules. This gives rise to semantic representations that are not tied to a particular modality but can be triggered by any of the sensory channels.
As we argue throughout this paper, this makes it possible to flexibly obtain new combinations of pieces of knowledge, enabling a compositional advantage aligned with the needs of systematic generalization out-of-distribution.

\subsection{Attention as dynamic information flow.}
The GWT suggests a fleeting memory capacity in which only one consistent content can be dominant at any given moment, which suggests a sharper form of attention than the soft attention currently dominant in deep learning and described below.
Attention is about sequentially selecting what computation to perform on what quantities. Let us consider a machine translation task from English to French. To obtain a good translation generating the next French word, we normally focus especially on the ``right'' few words in the source English sentence that may be relevant to do the translation. This is the motivation that stimulated our work on content-based soft self-attention~\citep{bahdanau2014neural}
but may also be at the heart of conscious processing in humans as well as in future deep learning systems with both system 1 and system 2 abilities.

%\vspace{-2mm}
\paragraph{Content-Based Soft Attention.} Soft attention forms a soft selection of one element (or multiple elements) from a set of elements at the previous level of computations, i.e we are taking a convex combination of the values of the elements at the previous level. These convex weights are coming from a softmax that is conditioned on how each of the elements' key vector matches some query vector.  In a way, attention is parallel, because computing these attention weights considers all the possible elements in some set, yielding a score for each of them, to decide which of them are going to receive the most attention. With stochastic hard-attention~\citep{xu2015show} one samples from a distribution over elements to choose the attended content, whereas with soft attention~\citep{bahdanau2014neural} one mixes these contents with different positive convex weights.  Content-based attention also introduces a non-local inductive bias into  neural network processing, allowing it to infer long-range dependencies that might be difficult to discern if computations are biased by local proximity. Attention is at the heart of the current state-of-the art NLP systems~\citep{DBLP:journals/corr/abs-1810-04805, brown2020language} and is the standard tool for memory-augmented neural networks \citep{graves2014neural,Sukhbaatar2015,gulcehre2016dynamic,santoro2018relational}.  Attention and memory can also help address the problem of credit assignment through long-term dependencies~\citep{ke2018sparse, kerg2020untangling} by creating dynamic skip connections through time (i.e., a memory access) which unlock the problems of vanishing gradients and learning long-term dependencies~\citep{hochreiter1991,Bengio:1994do}. Attention also transforms neural networks from machines that are processing vectors (e.g., each layer of a deep net), to machines that are processing sets, more particularly sets of key/value pairs, as with Transformers~\citep{vaswani2017attention,santoro2018relational}.
Soft attention uses the product of a \emph{query} (or \emph{read key}) represented as a matrix $Q$ of dimensionality $N_r \times d$, with $d$ the dimension of each key, with a set of $N_o$ objects each associated with a \emph{key} (or \emph{write-key}) as a row in matrix $K^T$ ($N_o \times d$), and after normalization with a softmax yields outputs in the convex hull of the \emph{values} (or \emph{write-values}) $V_i$ (row $i$ of matrix $V$). 
The result is 
$$\mathrm{Attention}(Q,K,V)=\mathrm{softmax} \left (\frac{QK^T}{\sqrt{d}} \right )V,$$
where the softmax is applied to each row of its argument matrix, yielding a set of convex weights. With soft attention, one obtains a convex combination of the values in the rows of $V$, whereas stochastic hard attention would sample one of the value vectors with probability equal to that weight. If the soft attention is focused on one element for a particular row (i.e., the softmax is saturated), we get deterministic hard attention: only one of the objects is selected and its value copied to row $j$ of the result.  Note that the $d$ dimensions in the key can be split into \emph{heads} which then have their attention matrix and write values computed separately. Note that hard attention is more biologically plausible (we only see one interpretation of the Necker cube~\citep{cohen1959rate} at once, and have one thought at a time) but soft attention enables end-to-end training and has been
the most commonly used in deep learning architectures up to now,. e.g., with transformers~\citep{vaswani2017attention}. However, there is recent evidence~\citep{liu2021discrete} that if the communication bottleneck is discretized, better OOD generalization is observed, maybe because the resulting simpler lingua franca would make it easier to swap one module for another in the attention-controlled communication between modules. 

%\vspace{-2mm}
\paragraph{Attention as dynamic connections.} We can think of attention as a way to create a dynamic connection between different blocks of computation, whereas in the traditional neural net setting, connections are fixed. On the receiving end (downstream module) of an attention-selected input, it is difficult to tell from the selected value vector from where it comes (among the selected upstream modules which competed for attention). To resolve this, it would make sense that the information being propagated along with the selected value includes a notion of key or type or name, i.e., of where the information comes from, hence creating a form of indirection (a reference
to where the information came from, which can be passed to downstream
computations). 

%\vspace{-2mm}
\paragraph{Attention implements variable binding.}
When the inputs and outputs of each of the modules are a set of objects or entities (each associated with a key and value vector), we have a generic object-processing machine which can operate on ``variables'' in a sense analogous to variables in a programming language: as interchangeable arguments of functions.
Because each object has a key embedding (which one can understand both as a name and as a type), the same computation can be applied to any variable which fits an expected ``distributed type'' (specified by a query vector).
Each attention head then corresponds to a typed argument of the function computed by the factor. When the key of an object matches the query of head $k$, it can be used as the $k$-th input vector argument for the desired computation.
Whereas in regular neural networks (without attention) neurons operate on fixed input variables (the neurons which are feeding them from the previous layer), the key-value attention mechanisms make it possible to select on the fly which variable instance (i.e. which entity or object) is going to be used as input for each of the arguments of some computation, with a different set of query embeddings for each argument head. 
The computations performed on the selected
inputs can be seen as \textit{functions with typed arguments}, 
and attention is used to \emph{bind} their formal argument to the selected input, albeit in a soft differentiable way (that mixes multiple possibilities) in the case of soft attention. Type constraints have already been found useful in identification for causal discovery~\citep{brouillard2022typing}. Current attention-based neural network already implement key-value-query soft attention mechanism (as above). What is missing is an ability to handle discrete types, hard (but possibly stochastic) choices of arguments, and more powerful inference machinery that uses not just type matching but is also able to reason about which modules and variables should be composed in a given context.

%\vspace{-2mm}
\subsection{Blend of Serial and Parallel Computations.} 

From a computational perspective, one hypothesis about the dynamics of communication between different modules is that different modules generally act in parallel and receive inputs from other modules. However, when they do need to communicate information with 
another {\em arbitrary} module, the information has to go through a routing bottleneck (the global workspace) controlled by an attention mechanism. Because so few elements can be put in coherence at each step of the GWT selection, the inference process generally requires several such steps, leading to the highly sequential nature of system 2 computation (compared with the highly parallel nature of system 1 computation). The contents which have thus been selected are essentially the only ones which can be committed to memory, starting with short-term memory. Working memory refers to the ability of the brain to operate on a few recently accessed elements (i.e., those in short-term memory) \citep{baddeley1992working, cowan1999embedded}. These elements can be remembered and have a heavy influence on the next thought, action or perception, as well as on what learning focuses on, possibly playing a role similar to desired outputs, goals or targets in supervised learning for system 1 computations.

\paragraph{Partial State.} From an RL perspective, it is interesting to note that if the GWT holds an important part of the state (including imagined future states, when planning), it does not describe all the aspects of the environment, only a handful of them, as already explored in the RL litterature~\citep{zhao2021consciousness}. This is different from standard RL approaches where the input (or the sequence of past inputs) is mapped to a fixed-size (estimated and latent) state vector. The GWT suggests instead that, besides long-term memory content (which mostly does not change), the rapidly changing state should be seen as a very small set of entities (e.g., objects or particular attributes of objects, and their relation), with an information content similar to that of a single sentence.  This suggests neural net architectures
in which very few modules and specific
(variable, value) pairs are selected at every inference step, based on those that were
recently selected, the current sensory input
and the current contents of memory (which can
also compete for write-access to the workspace).
Only the selected modules would be under pressure to adapt when the result of the combination needs to be tuned,
leading to selective adaptation similar to that explored by~\citep{bengio2019meta} (see Section~\ref{sec:meta-learning} above)
where just a few relevant modules need to
adapt to a change in distribution.

\paragraph{System 2 to System 1 Consolidation.} As an agent, a human being is facing frequent changes because of their actions or the actions of other agents in the environment. Most of the time, humans follow their habitual policy, but tend to use system 2 cognition when having to deal with unfamiliar settings. It allows humans to generalize out-of-distribution in surprisingly powerful ways, and understanding this style of processing would help us build these abilities in AI as well. This is illustrated with our early example of driving in an area with unfamiliar traffic regulations, which requires full conscious attention (Section~\ref{sec:conscious-vs-unconscious}). This observation suggests that system 2 cognition is crucial in order to achieve the kind of flexibility and robustness to changes in distribution required in the natural world \citep{shenhav2017toward, kool2018mental}. 
It looks like current deep learning systems are fairly good at perception and system 1 tasks. They can rapidly
produce an answer (if you have parallel computing like that of GPUs) through a complex calculation which is difficult
(or impossible) to dissect into the application of a few simple verbalizable operations. They require a lot of practice
to learn and can become razor sharp good at the kinds of data they are trained on. On the other hand, humans enjoy
system 2 abilities which permit fast learning (I can tell you a new rule in one sentence and you do not have to practice
it in order to be able to apply it, albeit awkwardly and slowly at first) and systematic generalization, both of which should be important characteristics of the next generation of deep learning systems.
 
\paragraph{Between-Modules Interlingua and Communication Topology.}
If the brain is composed of different modules, it is interesting to think about what code or lingua franca is used to communicate between them, such that it can lead to interchangeable pieces of knowledge being dynamically selected and combined to solve a new problem. The GWT bottleneck may thus also play a role in forcing the emergence of such a lingua franca~\citep{baars1997theatre, koch2004quest, shanahan2006cognitive}: the same information received by module A (e.g. ``there is a fire'') can come from any other module (say B, which detected a fire by smell,  or C which detected a fire by sight). Hence B and C need to use a compatible representation which is broadcast via the GWT bottleneck for A's use. Again, we see the crucial importance of attention mechanisms to force the emergence of shared representations and indirect references exchanged between the modules via the conscious bottleneck.
However, the GWT bottleneck is by far not the only way for modules to  communicate with each other. Regarding the topology of the communication channels between modules,  it is known that modules in the brain satisfy some spatial topology  such that the computation is not all-to-all between all the modules. It is plausible that the brain uses both fixed local or spatially nearby connections  as well as the global broadcasting system with top-down influence. We also know  that there are hierarchical communication routes in the visual cortex (on the path from pixels to object recognition), and we know how successful that has been in computer vision with convnets. Combining these different kinds of inter-module communication modalities in deep network thus
seems well advised as well \citep{watts1998collective, latora2001efficient,rahaman2020s2rms}: 
(1) Modules which are near each other in the brain layout can probably communicate directly without the need to clog the global broadcast channel (and this would not be reportable consciously). 
(2) Modules which are arbitrarily far from each other in the spatial layout of the brain can exchange information via the global workspace, following the theatre analogy of Baars' GTW. The other advantage of this communication route is
    of course the exchangeability of the sources of information being broadcast, which we hypothesize leads to
    better systematic generalization.
The role of working memory in the GWT is not just as a communication
buffer. It also serves as a blackboard (or analogously the ``registers'' in CPUs) where operations can be done locally to improve coherence. This enables a coherence-seeking mechanism: the different modules (especially the active ones) should adopt a configuration of their internal variables (and especially the more abstract entities they communicate to other modules) which is consistent with what other active modules ``believe''. It is possible, that a large part of the functional role of conscious processing is for that purpose, which is consistent with the view of the working memory as a central element of the inference machinery seeking to obtain coherent configurations of the variables interacting according to some piece of knowledge (such as a factor of the factor graph, a causal dependency).
 
\begin{myremark}{System 2 inductive biases}
 We are proposing to take inspiration from cognition and build machines which integrate two very different kinds of representations and computations corresponding to the system 1 / implicit / unconscious vs system 2 / explicit / conscious divide.
\end{myremark}

This paper is about inductive biases not yet sufficiently integrated in state-of-the-art deep learning systems but which could help us achieve these system 2 abilities. In the next subsection, we summarize some of these system 2 inductive biases.

\subsection{Semantic Representations Describing Verbalizable Concepts}
\label{sec:semantic=high-level}

%\Ani{abrupt starting to this paragraph, improve it.}
Conscious content is revealed by reporting it, often with language \citep{colagrosso2004theories}. This suggests that high-level variables manipulated consciously are closely related with their verbal forms (like words and phrases). This yields maybe the most influential inductive bias we want to consider in this paper: that \textit{high-level variables (manipulated consciously) are generally verbalizable}. To put it in simple terms, we can imagine the high-level
semantic variables captured at this top level of a representation
to be associated with single words (although we can also use words to identify some lower-level variables). In practice, the notion of word is not always the same across different languages, and the same semantic concept may be represented by a single word or by a phrase. There may also be more subtlety in the mental representations
(such as accounting for uncertainty, concept representation and continuous-valued properties) 
which is not always or not easily well reflected in their verbal rendering.
Much of what our brains know actually cannot be easily translated in natural language and forms the content of system 1 knowledge. This means that system 2 (verbalizable) knowledge is incomplete: words are mostly pointers to knowledge which belongs to system 1
and thus is in great part not consciously accessible.
The system 2 inductive biases do not need to cover all the aspects of our internal model of the world (they couldn't), only those aspects of our knowledge which we are able to communicate with language. The rest would have to be represented
in pure system 1 (non system 2) machinery, such as in an encoder-decoder that could relate low-level actions and low-level perception to semantic variables that can be operated on at the system-2 level.  If there is some set of properties that apply 
well to some aspects of the world, then it would be advantageous for a learner to have a subsystem that takes advantage of these properties (the inductive priors described here) and a subsystem which models the other aspects. These inductive priors then allow faster learning and potentially other advantages like systematic generalization, at least concerning these aspects of the world which are consistent with these assumptions (system 2 knowledge, in our case).

%The assumption we are making is thus that

\begin{myremark}{High-level representations describe verbalizable concepts}
There is a simple lossy mapping from semantic representations going through the GWT bottleneck to natural language expressions. This is an inductive bias which could be exploited in grounded language learning scenarios~\citep{winograd1972understanding, hermann2017grounded, chevalier2018babyai, hill2019environmental} where we couple language data with observations and actions by an agent.
\end{myremark}
This suggests that natural language understanding systems should be trained in a way that couples natural language with what it refers to. This is the idea of {\em grounded language  learning}. It would put pressure on the top-level representation so that it captures the kinds of concepts expressed with language. One can view this as a form of weak supervision, where we don't force the top-level GWT representations to be human-specified labels, only that there is a simple relationship between these representations and utterances which humans would often associate with the corresponding meaning.
Our discussion about causality should also suggest that passive observation may be insufficient: in order to capture the causal structure understood by humans, it may be necessary for learning agents to be embedded in an environment in which they can act and thus discover its causal structure \citep{binz2022using, kosoy2022towards}. Studying this kind of setup was the motivation for our work on the Baby AI environment~\citep{chevalier2018babyai}.

\vspace{-2mm}
\subsection{Semantic Variables Play a Causal Role and Knowledge about them is Modular}
\label{sec:semantic-variables-are-causal}

Biological phenomena such as bird flocks have inspired the design of several distributed multi-agent systems, for example, swarm robotic systems, sensor networks, and modular robots.  Despite this, most machine learning models employ the opposite inductive bias, i.e., with all elements (e.g., artificial neurons) interacting all the time. The GWT~\citep{baars1997theatre,dehaene2020we} also posits that the brain is composed in a modular way,
with a set of expert modules which need to communicate but only do so sparingly and via a bottleneck through which only a few selected bits of information can be squeezed at any time. If we believe that theory, these selected elements are the concepts present to our mind at any moment, and a few of them are called upon and joined in working memory in order to reconcile the interpretations made by different modular experts across the brain. The decomposition of knowledge into recomposable pieces, a hallmark of classical AI based on rules \citep{russell2010artificial} also makes sense as a requirement for obtaining systematic generalization \citep{bahdanau2018systematic}: conscious attention would then select which expert and which concepts (which we can think of as variables with different attributes and values) interact with which pieces of knowledge (which could be verbalizable rules or non-verbalizable intuitive knowledge about these variables) stored in the modular experts. On the other hand, the modules which are not brought to bear in this conscious processing may continue working in the background in
a form of default or habitual computation (which would be the form of most of perception). For example, consider the task of predicting from pixel-level information the motion of balls sometimes colliding against each other as well as the walls. It is interesting to note that all the balls follow their default dynamics, and only when balls collide do we need to intersect information from several bouncing balls in order to make an inference about their future states.  Saying that the brain modularizes knowledge is not sufficient, since there could be a huge number of ways of factorizing knowledge in a modular way. We need to think about the desired properties of modular decompositions of the acquired knowledge, and we propose here to take inspiration from the causal perspective on understanding how the world works, to help us define both the right set of variables and their relationship. 

\begin{myremark}{Semantic variables are often also causal variables}
We hypothesize that semantic variables are often also causal variables. Words in natural language often refer to agents (subjects, which cause things to happen), objects (which are controlled by agents),  actions (often through verbs) and modalities or properties of agents, objects and actions (for example we can talk about future actions, as intentions, or we can talk about time and space where events happen, or properties of objects or of actions).  However, note that we can also name many low-level (like pixels) and intermediate features (like L-shaped edges). It is thus plausible to assume that causal reasoning of the kind we can verbalize involves as variables of interest those semantic variables which we can name, and that they can be at any level of the processing hierarchy in the brain, including at the highest levels of abstraction, where signals from all modalities join, such as pre-frontal cortex~\citep{cohen2000anterior}, and where concepts
can be manipulated in a way that is not specific to a single modality.
\end{myremark}

The connection between causal representations and modularity is profound: an assumption which is commonly associated with structural causal models is that it should break down knowledge about the causal influences into {\em independent mechanisms}~\citep{peters2017elements}.  As explained in Section~\ref{sec:icm}, each such mechanism relates direct causes to their direct effect and knowledge of one such mechanism should not tell us anything about another mechanism (otherwise we should restructure our representations and decomposition of knowledge to satisfy this information-theoretic independence property).
This is not about statistical independence of the corresponding random variables but about the algorithmic mutual information between the descriptions of these mechanisms. What it means practically and importantly for out-of-distribution adaptation is that if a mechanism changes (e.g. because of an intervention), the representation of that mechanism (e.g. the parameters used to capture a corresponding conditional distribution) may
need to be adapted but that of the others do not need to be tuned to account for that change~\citep{bengio2019meta}.  

These mechanisms may be organized in the form of a causal graph which scientists attempts to identify.  The sparsity of the change in the joint distribution between the semantic variables (discussed more in Section~\ref{sec:sparse-change}) is different but related to a  property of such high-level structural causal model: the sparsity of the graph capturing the joint distribution itself
(discussed in Section~\ref{sec:sparse-graph}). In addition, the causal structure, the causal mechanisms and the definition of the high-level causal variables  tend to be stable across changes in distribution, as discussed in Section~\ref{sec:stable}. \newline

\vspace{-4mm}
\subsection{Local Changes in Distribution in Semantic Space}
\label{sec:sparse-change}

Consider a learning agent, like a learning robot or a learning child. What are the sources of non-stationarity
for the distribution of observations seen by such an agent, assuming the environment is in some (generally
unobserved) state at any particular moment? 
Two main sources are (1) the non-stationarity due to the environmental dynamics (including the learner's actions and policy) not having
converged to an equilibrium distribution (or equivalently the mixing time of the environment's stochastic
dynamics is longer than the lifetime of the
learning agent) and (2) causal interventions by agents (either the learner of interest or some other agents).
The first type of change includes for example the case of a person moving to a different country, or a videogame
player learning to play a new game or a never-seen level of an existing game. That first type also includes the non-stationarity
due to changes in the agent's policy arising from learning.
The second case includes the effect of actions such as locking some doors in a labyrinth (which may have a drastic effect on the optimal policy). The two types can intersect, as the actions of agents (including those of the learner, like
moving from one place to another) contribute to the first type of non-stationarity.

\begin{myremark}{Changes in distribution are localized in the appropriate semantic space}
Let us consider how humans describe these changes with language. For many of these changes, they are able to explain
the source of change with a few words (a single sentence, often). This is a very strong clue for our proposal to include as an inductive bias the assumption that \textit{ the source of most changes in distribution are localized in the appropriate semantic space}: only one or a few variables or mechanisms need to be modified to account for the change.
\end{myremark}

Note how humans will even create new words when they are not able to explain a change with a few existing words, with the new words corresponding to new latent variables, which when introduced, make the changes explainable ``easily'' (assuming one understand the definition of these variables and of the mechanisms relating them to other variables). 

For system-2 distributional changes (due to interventions), we automatically get locality of the source of changes (which start at one or a few nodes of the causal graph). This is a plausible assumption since, by virtue of being localized in time and space, actions can only directly affect very few high-level variables, with other effects (on downstream variables) being consequences of the initial intervention. This sparsity of sources of change is a strong assumption which can put pressure on the learning process to discover high-level representations which have that property. Here, we are assuming that the learner has to jointly discover these high-level representations (i.e. how they relate to low-level observations
and low-level actions) as well as how the high-level variables relate to each other via causal mechanisms. 

\vspace{-4mm}
\subsection{Stable Properties of the World}
\label{sec:stable}

Above, we have talked about changes in distribution due to non-stationarities, but there are aspects of the world that are stationary, which
means that learning about them would eventually converge. In an ideal scenario, our learner has an infinite lifetime
and the chance to learn
everything about the world (a world where there are no other agents)
and build a perfect model of it, at which point nothing is new and all of the above
sources of non-stationarity are gone. In practice, only a small part of the world will be
understood by the learning agent, and interactions between agents (especially if they are learning) will
perpetually keep the world out of equilibrium. If we divide the knowledge about the world
captured by the agent into the stationary aspects (which should converge) and the non-stationary
aspects (which would generally keep changing), we would like to have as much knowledge as possible
in the stationary category. The stationary part of the model might require many observations for
it to converge, which is fine because learning these parts can be amortized over the whole
lifetime (or even multiple lifetimes in the case of multiple cooperating cultural agents, e.g., in human
societies). On the other hand, the learner should be able to quickly learn the non-stationary parts 
(or those the learner has not yet realized can
be incorporated in the stationary parts), ideally because very few of these parts need to change, if knowledge
is well structured. Hence we see the need for at least two speeds of learning, similar to the
division found in meta-learning
of learnable coefficients into meta-parameters on one hand (for the stable, slowly learned aspects)
and parameters on the other hand (for the non-stationary, fast to learn aspects), as already discussed above
in Section~\ref{sec:about-inductive-biases}. 

%The proposed inductive bias (which now involves the whole system, not just system 2), is that t

\begin{myremark}{Stable v/s Unstable properties of the world}
There should be several speeds of learning, with more stable aspects learned more slowly and more non-stationary or novel ones learned faster, and pressure to discover stable aspects among the quickly changing ones. This pressure would mean that more aspects of the agent's represented knowledge of the world become stable and thus less needs to be adapted when there are changes in distribution. 
\end{myremark}

For example, consider scientific laws, which are most powerful when they are universal.
At another level, consider the mapping between the perceptual input, low level actions, and the high-level semantic variables. 
An encoder that would implement this mapping should ideally be highly stable, or else downstream computations would need to
track those changes (and indeed the low-level visual cortex seems to compute features
that are very stable across life, contrary to high-level concepts like new visual categories). 
Causal interventions are taking place at a higher level than the encoder, changing the value of an unobserved high-level variable or changing one of the
mechanisms. If a new concept is needed, it can be added without having to disturb other represented knowledge, especially if it can be
learned as a composition of existing high-level features and concepts.
We know from observing humans and their brain that new concepts which are not obtained from
a combination of old concepts (like a new skill or a completely new object category not obtained
by composing existing features) take more time to learn, while new high-level concepts which can be readily
defined from other high-level concepts can be learned very quickly (as fast as with a single example or definition). 

Another example arising from the analysis of causal systems is that causal interventions (which are in the non-stationary, quickly inferred or quickly learned category) may temporarily modify the causal graph structure
(which specifies which variable is a direct cause of which) by breaking causal links (when we set a variable we break the causal link from its direct causes) but that most of the causal graph is a stable property of the environment. Hence, we need neural architectures which make it easy to quickly adapt the relationship between existing concepts, or to define new concepts from existing ones.

\vspace{-4mm}
\subsection{Sparse Factor Graph in the Space of Semantic Variables}
\label{sec:sparse-graph}

\begin{myremark}{Sparsity as to how variables and factors interact with each other}
Our next inductive bias for high-level variables can be stated
simply: the joint distribution between high-level concepts  can be represented by a sparse factor graph.
\end{myremark}

Any joint distribution can be expressed as a factor graph \citep{kschischang2001factor, frey2012extending, kok2005learning}, but we claim that the ones which can be conveniently described with natural language have the property that they should be sparse. A factor graph is a particular factorization of the joint distribution. A factor graph is bipartite, with variable nodes on one hand and factor nodes on the other. Factor nodes represent dependencies between the variables to which they are connected. To illustrate the sparsity of verbalizable knowledge, consider knowledge graphs and other relational systems, in which relations  between variables often involve only two arguments (i.e., two variables). In practice, we may want factors with more than two arguments, but probably not a lot more. A factor may capture a causal mechanism between its argument variables, and thus we should introduce an additional semantic element to these factors: each argument of a causal factor should either play the role of cause or of effect, making the bipartite graph directed.
 
It is easy to see that linguistically expressed knowledge satisfies this sparsity property by noting that statements about the world can be expressed with a sentence and each sentence typically has only a few words, and thus relates very few concepts. When we write ``If I drop the ball, it will fall on the ground'', the sentence clearly involves very few variables, and  yet it can make a very strong prediction about the position of the ball. A factor in a factor graph involving a subset $S$ of variables is simply stating a probabilistic constraint among these variables. It allows one to predict the value of one variable given the others (if we ignore other constraints or factors), or more generally it allows us to describe a preference for joint sets of values for a subset of $S$. The fact that natural language allows us to make such strong predictions conditioned on so few variables should be seen as surprising: it only works because the variables are semantic ones. If we consider the space of pixel values in images, it is very difficult to find such strongly predictive rules, e.g., to predict the value of one pixel given the value of three other pixels. What this means is that pixel space does not satisfy the sparsity prior associated with the proposed inductive bias.

We claim that the proposed inductive bias is closely related to the bottleneck of the GWT of conscious processing. Our interpretation of this restriction on write access in the GWT by a very small number of specialists selected on the fly by an attention mechanism is that it stems from an  assumption on the form of the joint distribution between high-level variables whose values are broadcast.
If the joint distribution factor graph is sparse, then only a few variables (those involved in one factor or a few connected factors) need to be synchronized at each step of an inference process, e.g., consider loopy belief propagation~\citep{frey2001very,murphy2013loopy}. By constraining the size of the working memory, evolution may have thus enforced the sparsity of the factor graph. The GWT also makes a claim that the workspace is associated  with the conscious contents of cognition, which can be reported verbally. One can also make links with the original von Neumann architecture of computers. In both the GWT and
the von Neumann architecture, we have a communication bottleneck with in the former the working memory and
in the latter the CPU registers where operations are performed. The communication bottleneck only allows
a few variables to be brought to the nexus (working memory in brains, registers in the CPU) . In addition, the operations on these variables are extremely sparse,
in the sense that they take very few variables at a time as arguments (no more than the handful in working memory,
in the case of brains, and generally no more than two or three in typical assembly languages). This sparsity
constraint is consistent with a decomposition of computation in small chunks, each involving only
a few elements. In the case of the sparse factor graph assumption we only consider that sparsity constraint
for declarative knowledge (verbalizing "how the world works", its dynamics and statistical or causal structure).

This assumption about the joint distribution between the high-level variables at the top of our deep learning hierarchy is different from the assumption commonly found in many papers on disentangling factors of variation~\citep{higgins2016beta, burgess2018understanding, chen2018isolating, kim2018disentangling, locatello2019challenging}, where the high-level variables are assumed to be marginally independent of each other,
i.e., their joint distribution factorizes into independent marginals. We think this deviates from the original goals of deep learning to learn abstract high-level representations which capture the underlying explanations for the data. Note that one can easily transform one representation (with a factorized joint) into another (with a non-factorized joint) by some transformation (think about the independent noise variables in a structural
causal model, Section~\ref{sec:scm}). However, we would then lose the properties introduced up to now (that each variable is causal and corresponds to a word or phrase, that the factor graph is sparse, and that changes in distribution can be originated to one or very few variables or factors). 

Instead of thinking about the high-level variables as completely independent, we propose to see them as having a very structured joint distribution, with a sparse factor graph and other characteristics (such as dependencies which can be instantiated on particular variables from generic schemas or rules, described below). We argue that if these high-level variables have to capture semantic variables expressible with natural language,  then the joint distribution of these high-level semantic variables  must have sparse dependencies rather than being independent. For example, high-level concepts such as "table" and "chair" are not statistically independent, instead they come in very powerful and strong but sparse relationships. Instead of imposing a very strong prior of complete independence at the highest level of representation, we can have this slightly weaker but very structured prior, that the joint is represented by a sparse factor graph. Interestingly, recent studies confirm that the top-level variables in generative adversarial networks (GANs), which are independent by construction, generally do not have a semantic interpretation (as a word or short phrase), whereas many units in slightly lower layers do
have a semantic interpretation~\citep{bau2018gan}.

Why not represent the causal structure with a directed graphical model? In these models, which are the basis of
standard representations of causal structure (e.g., in structural causal models, described below), knowledge to be learned is
stored in the conditional distribution of each variable (given its direct causal parents). However, it is not clear that this
is consistent with the requirements of independent mechanisms. For example, typical verbally expressed rules have the property
that many rules could apply to the same variable. Insisting that the independent units of knowledge are conditionals would then necessarily
lump the corresponding factors in the same conditional. This issue becomes even more severe if we think of the rules as
generic pieces of knowledge which can be reused to be applied to many different tuples of instances, as elaborated in
the next subsection. Another reason for a formulation that is not constrained to an acyclic graph is that humans also reason about relations between variables at equilibrium (such as voltage and current), which can mutually be causes of each other (i.e., arrows can go both ways).

\vspace{-4mm}
\subsection{Variables, Instances and Reusable Knowledge Pieces}
\label{sec:generic-knowledge}
\label{sec:inference-vs-declarative}

A standard graphical model is static, with a separate set of
parameters for each conditional distribution (in a directed acyclic graph) or factor (in a factor graph). There are extensions which allow parameter sharing, e.g. through time with dynamic Bayes nets~\citep{spirtes2000causation},  or in undirected graphical models such as Markov Networks~\citep{kok2005learning} which allow one to ``instantiate'' general ``patterns'' into multiple factors of the factor graph. Markov Networks can for example implement forms of recursively applied probabilistic rules. But they do not take advantage of distributed representations and other inductive biases of deep learning.

The inductive bias we are presenting here is that instead of separately defining specific factors in the factor graph (maybe each with a piece of neural network), each having its separate set of parameters, we would define generic factors, ``schemas'' or  ``factor templates''. A schema, or generic factor is a reusable probabilistic relation, i.e., with argument variables which can be bound to instances (also discussed in ~\citep{rumelhart1986sequential}). A static instantiated rule is a thing like 'if John is hungry then he looks for food'. Instead, a more general rule is a thing like, 'for all $X$, if $X$ is a human and $X$ is hungry, then $X$ looks for food' (with some probability). $X$ can be bound to specific instances (or to other variables which may involve more constraints on the acceptable set). In classical symbolic AI, we have unification mechanisms to match together variables, instances or expressions involving variables and instances, and thus keep track of how variables can ultimately be 'bound' to instances (or to variables with more constraints on their
attributes), when exploring whether some schema can be applied to some objects (instances or more generic objects) with properties (constituting a database of entities).

The proposed inductive bias is also inspired by the presence of such a structure in the semantics of natural language and the way we tend to organize knowledge according to relations, e.g., in knowledge graphs~\citep{sowa1987semantic}. Natural language allows us to state rules involving variables and is not limited to making statements about specific instances. 

\begin{myremark}{Knowledge is generic and can be instantiated on different instances.}
The independent mechanisms (with separate parameters) which specify dependencies between variables are generic, i.e., they can be instantiated in many possible ways to specific sets of arguments with the appropriate types or constraints.
\end{myremark}

What this means in practice is that we do not need to hold in memory the full instantiated graph with all possible instances
and all possible mechanisms relating them (or worse, all the
generic factor instantiations that are compatible with the data, in a Bayesian posterior). Instead, inference involves generating the needed pieces of the graph and  even performing reasoning (i.e. deduction) at an abstract level, where nodes in the graph
(random variables) stand not for instances but for sets of instances belonging to some category or satisfying
some constraints. Whereas one can unfold a recurrent neural network or a Bayesian network
to obtain the fully instantiated graph, in the case we are talking about, similarly to a Markov network,
it is generally not feasible to do that. It means that inference procedures always look at a small piece of the (partially) unfolded graph at a time and they can reason about how to combine these generic schemas without having to fully instantiate them with
concrete instances  or concrete objects in the world. One way to think about this, inspired by how we do programming, is that we have functions with generic and possibly typed variables as arguments and we have instances on which a program is going to be applied. At any time (as you would have in Prolog), an inference engine must match the rules with the current instances (so the types and other constraints between arguments are respected) as well as other elements (such as what we are trying to achieve
with this computation) in order to combine the appropriate computations. It would make sense to think of such a computation
controller, as an internal policy with attention and memory access as actions, to select which pieces of knowledge and which pieces of the short-term (and occasionally long-term) memory need to be combined in order to push new values in working memory \citep{shanahan2005applying, shanahan2006cognitive, baars1993cognitive, baars1997theatre}.

An interesting outcome of such a representation is that one can apply the same knowledge (i.e knowledge specified by a schema which links multiple abstract entities together) to different instances (i.e different ``object files'' in cognitive psychology \citep{noles2005persistence, gordon1996s, kahneman1992reviewing}). For example, you can apply the same laws of physics to two different balls that are visually different (and maybe have different colors and masses). This is also related to notions of arguments and indirection in programming.  The  power of such relational reasoning  resides  in  its  capacity  to generate inferences and generalizations that are constrained by the roles that elements play, and the roles they can play may depend on the properties of these elements, but these schemas specify how entities can be related to each other in systematic (and possibly novel) ways. In the limit, relational reasoning yields universal inductive generalization from a finite and often very small set of observed cases to a potentially infinite set of novel instances, so long as those instances can be described by attributes (specifying types) allowing to bound them to appropriate schemas.

%\paragraph{Inference versus declarative knowledge}

There are two forms of knowledge representation we have discussed: 
declarative knowledge or hypotheses, i.e., that can be verbalized (e.g. of facts, hypotheses, explicit causal dependencies, etc), and inference machinery used to reason with these pieces of knowledge. Standard graphical models only represent the declarative knowledge and typically require expensive but generic iterative computations (such as Monte-Carlo Markov chains) to perform approximate
inference \citep{cowles1996markov, gilks1995markov}. However, brains need fast inference \citep{gigerenzer1996reasoning}, and most of the advances made with deep learning concern such learned fast inference computations. Doing inference using only the declarative knowledge (the graphical model) is very flexible (any question of the form ``predict some variables given other variables or imagined interventions'' can be answered) but also very slow.
In general, searching for a good configuration of the values of top-level variables which is consistent with the given context is computationally intractable. However, different approximations can be made which trade-off computational cost for quality of the solutions found. This difference could also be an important ingredient of the difference between system 1 (fast and parallel approximate and inflexible inference) and system 2 (slower and sequential but more flexible inference). We also know that after system 2 has been called upon to deal with novel situations repeatedly, the brain tends to bake these patterns of response in habitual system 1 circuits which can do the inference job faster and more accurately but have lost some flexibility. When a new rule is introduced, the system 2 is flexible enough to handle it and slow inference needs to be called upon again. Neuroscientists have also accumulated evidence that the hippocampus is involved in replaying sequences (from memory or imagination) for consolidation into cortex \citep{alvarez1994memory, hassabis2007using} so that they can be presumably committed to cortical long-term memory and fast inference.

\vspace{-2mm}
\subsection{Relevant causal chains (for learning or inference) can be approximated as very short chains}
\label{sec:short-chains}

% BRAIN statement
In a \textit{clock-based segmentation}, the boundaries between discrete time steps are spaced equally \citep{hihi1995hierarchical, chung2016hierarchical, koutnik2014clockwork}. In an \textit{event-based segmentation}, the boundaries depend on the state of the environment, resulting in dynamic duration of intervals \citep{mozer1997parsing}. 
Our brains seem to segment streams of sensory inputs into meaningful representations of variable-length episodes and \emph{events} \citep{suddendorf2007evolution, ciaramelli2008top, berntsen2013remembering, dreyfus1985socrates, richmond2017constructing}. 

The detection of a relevant event in the temporal stream triggers information processing of the event. The psychological reality of event-based segmentation can be illustrated through a familiar phenomenon. Consider the experience of traveling from one location to another, such as from home to office. If the route is unfamiliar, as when one first starts a new job, the trip is confusing and lengthy, but as one gains more experience following the route, one has the sense that the trip becomes shorter. One explanation for this phenomenon is as follows. On an unfamiliar route, the orienting mechanism that detects novel events is triggered for a large number of such events over the course of the trip. In contrast, few novel events occur on a familiar route. If our perception of time is event-based, meaning that higher centers of cognition count the number of events occurring in a temporal window, not the number of milliseconds, then one will have the sense that a familiar trip is shorter than an unfamiliar trip.

Event segmentation allows functional representations that support temporal reasoning, an ability that arguably relies on neural circuits to encode and retrieve information to and from memory \citep{zacks2007event, radvansky2017event, baldassano2017discovering}. Indeed, faced with a task, our brains appear to easily and {\it selectively} pluck context-relevant past information from memory, enabling both powerful multi-scale associations as well as flexible computations to relate temporally distant events. As we argue here, the ability of the brain to efficiently segment sensory inputs into events, and the ability to \emph{selectively} recall information from the distant past based on the current context helps to efficiently propagate information (such as credit assignment or causal dependencies) over long time spans.   Both at the cognitive and at the physiological levels, there is evidence of information ``routing'' mechanisms that enable this efficient propagation of information, although they are far from being sufficiently understood \citep{stocco2010conditional, ben2018hippocampal, bonasia2018prior}.

\begin{myremark}{Relevant Causal Chains tend to be sparse.}
Our next inductive bias is almost a consequence of the biases on causal variables and the bias on the sparsity of
the factor graph for the joint distribution between high-level variables. Causal chains used to perform learning (to imagine counterfactuals and to propagate and assign credit) or inference (to obtain explanations or plans for achieving some goal) are broken down into short causal chains of events which may be far in time but linked by the top-level factor graph over semantic variables.
\end{myremark}

At least at a conscious level, humans are not able to reason about many such events at a time, due to the limitations on short-term memory and the bottleneck of conscious processing~\cite{baars1997theatre}.  Hence it is plausible that humans would exploit an assumption on temporal dependencies in the data: that the most relevant ones only involve short dependency chains, or a small-depth graph of direct dependencies. Depth here refers to the longest path in the relevant graph of dependencies between events. What we showed
earlier~\citep{ke2018sparse, kerg2020untangling} is that this prior assumption is the strongest ingredient to mitigate the issue of vanishing gradients that occurs when trying to learn long-term dependencies~\citep{Bengio:1994do}.

\vspace{-2mm}
\subsection{Context-dependent processing involving goals, top-down influence, and bottom-up competition}
\label{sec:bottom-up-vs-top-down}

Successful perception in humans clearly relies on both top-down and bottom-up signals \citep{buschman2007top, beck2009top, mcmains2011interactions, kinchla1979order, rauss2013bottom, mcclelland1981interactive}.  Top-down information encodes relevant context, priors and preconceptions about the current scene: for example, what we might expect to see when we enter a familiar place.  Bottom-up signals consist of what is literally observed through sensation.  The best way to combine top-down and bottom-up signals remains an open question, but it is clear that these signals need to be combined in a way which is dynamic and depends on context - in particular top-down signals are especially important when stimuli are noisy or hard to interpret by themselves (for example walking into a dark room).  Additionally, which top-down signals are relevant also changes depending on the context. It is possible that combining specific top-down and bottom-up signals that can be weighted dynamically (for example using attention) could improve robustness to distractions and noisy data.

In addition to the general requirement of dynamically combining top-down and bottom-up signals, it makes sense to do so at every level of the processing hierarchy to make the best use of both sources of information at every stage of that computation, as is observed in the visual cortex (with very rich top-down signals influencing the
activity at every level). 

\begin{myremark}{Dynamic Integration of Bottom-up and Top-Down Information}
In favour of architectures in which top-down contextual information is dynamically combined with bottom-up sensory signals at every level of the hierarchy of computations relating low-level and high-level representations.
\end{myremark}

\section{Declarative Knowledge of Causal Structure}
\label{sec:scm}

Whereas a statistical model captures a single joint distribution, a causal model
captures a large family of joint distributions, each corresponding to a different intervention (or set of interventions), which modifies
the unperturbed or default distribution (e.g., by removing parents of a node and setting a value for that node). Whereas the
joint distribution $P(A,B)$ can be factored either as $P(A)P(B|A)$ or $P(B)P(A|B)$ (where in general both graph
structures can fit the data equally well), only one of the graphs corresponds to the correct causal structure and
can thus consistently predict the effect of interventions. The asymmetry is best illustrated by an example:
if $A$ is altitude and $B$ is average temperature, we can see that intervening on $A$ will change $B$ but not vice-versa.

\vspace{-2mm}
\paragraph{Preliminaries}
Given a set of random variables $X_i$, a Bayesian network is commonly used to describe the dependency structure of both probabilistic and causal models via a Directed Acyclic Graph (DAG). In this graph structure, a variable (represented by a particular node) is independent of all the other variables, given all the direct neighbors of a variable.  The edge direction identify a specific factorization of the joint distribution of the graph's variables:
\begin{equation}\label{eq:cf}
p(X_1,\dots,X_n) = \prod_{i=1}^m  p(X_i \mid \PA_i).
\end{equation}

\vspace{-2mm}
\paragraph{Structural causal models (SCMs).}
A Structural Causal Model (SCM)  \citep{peters2017elements} over a finite number $M$ of random variables $X_i$ given a set of \textit{observables} $X_1,\dots,X_M$ (modelled as random variables) associated with the vertices of a DAG $G$, is a set of structural assignments
\begin{align}\label{eq:SA}
	X_i &:= f_i(X_{pa(i,C)}, N_i)\;, \quad \forall i \in \{1,\ldots,M\}
\end{align}
where $f_i$ is a deterministic function,  the set of noises $N_1,\dots, N_m$ are assumed to be {\em jointly independent}, and $pa(i,C)$ is the set of parents (direct causes) of variable $i$ under configuration $C$ of the SCM directed acyclic graph, i.e., $C \in \{0,1\}^{M \times M}$, with $c_{ij} = 1$ if node $i$ has node $j$ as a parent (equivalently, $X_j \in X_{pa(i,C)}$; i.e. $X_j$ is a direct cause of $X_i$). Causal structure learning is the recovery of the ground-truth $C$ from observational and interventional data, possibly yielding
a posterior distribution over causal structures compatible
with the data, and a neural network can be trained to generate graphs from that posterior~\citep{deleu2022bayesian}.
%
%An SCM's induced graph has a strictly lower-triangular adjacency matrix $C$ if the nodes are sorted in topological order. 
%The $n$-th power of an adjacency matrix, $C^n$, counts the number of length-$n$ walks from node $i$ to node $j$ of the graph in element $c_{ij}$. The trace of the $n$-th power of an adjacency matrix, $\mathrm{Tr}(C^n)$, counts the number of length-$n$ cycles in the graph.
%Intuitively, we can think of the independent noises as ``information probes'' that spread through the graph (much like independent elements of gossip can spread through a social network). Their information gets entangled, manifesting itself in a footprint of conditional dependences rendering the possibility to infer aspects of the graph structure from observational data using independence testing. Like in the gossip analogy, the footprint may not be sufficiently characteristic to pin down a unique causal structure and one may need to perform experiments or interventions (or observe interventions) in order to fully recover the causal graph. 

\vspace{-2mm}
\paragraph{Interventions. } Without experiments, or interventions i.e., in a purely-observational setting, it is known that causal graphs can be distinguished only up to a Markov equivalence class,
i.e., the set of graphs compatible with the observed dependencies. In order to identify the true causal graph, the learner needs to perform interventions or experiments i.e., interventional data is generally needed \citep{eberhardt2012number}. 

\vspace{-4mm}
\subsection{Independent Causal Mechanisms\label{sec:icm}.} 

A powerful assumption about how the world works which arises from research in causality~\citep{peters2017elements} and briefly introduced earlier is that the causal structure
of the world can be described via the composition of independent causal mechanisms.
%Here, we will consider notion of independence among different factors in the factorization of join distribution in eq. \eq{eq:cf} relating one factor to another factor already introduced above in (Section~\ref{sec:semantic-variables-are-causal} and~\ref{sec:generic-knowledge}).

\hypertarget{pri:im}{\textit{ Independent Causal Mechanisms (ICM) Principle.}} \hspace{0.1mm}
{\em A complex generative model, temporal or not, can be thought of as composed of independent mechanisms that do not inform or influence each other. In the probabilistic case, this means a particular mechanism should not inform (in the information theory sense) or influence the other mechanisms. }

This principle subsumes several notions important to causality, including separate intervenability of causal variables, modularity and autonomy of subsystems, and invariance \citep{Pearl2009,PetJanSch17}.

This principle applied to the factorization in equation. \ref{eq:cf},  tells us that the different factors should be independent in the sense that (a) performing an intervention on one of the mechanisms $p(X_i|\PA_i)$ does not change any of the other mechanisms $p(X_j|\PA_j)$ ($i\ne j$), (b) knowing some other mechanisms $p(X_i|\PA_i)$ ($i\ne j$) does not give us information about any another mechanism $p(X_j|\PA_j)$.

\vspace{-2mm}
\subsection{Exploit changes in distribution due to causal interventions}

\paragraph{Nature doesn't shuffle examples.} Real data arrives to us in a form which is not iid, and so in practice what many practitioners of data science or researchers do when they collect data is to \textit{shuffle} it to make it iid. ``Nature doesn't shuffle data, and we should not'' \cite{bottouiclrtalk}. When we shuffle the data, we destroy useful information about those changes in distribution that are inherent in the data we collect and contain information about causal structure. Instead of destroying that information about non-stationarities, we should use it, in order to learn how the world changes.

%\paragraph{Challenges for Deep Learning.} Most work in causality starts from the premise that the causal variables themselves have known semantics or are observed. However, for AI agents such as robots trying to make sense of their environment, the only observables are low-level variables like pixels in images or low-level motor actions. To generalize well, an agent must induce high-level variables, particularly those which are causal (i.e., can play the role of cause or effect). A central goal for AI and causality research is thus the joint discovery of abstract high-level representations (how they relate to low-level observations and low-level actions) at the same time as of causal structure at the high level. This objective also dovetails with the goals of the representation learning and structure learning communities.

\subsection{Relation between meta-learning, causality, OOD generalization and fast transfer learning}
\label{sec:meta-learning}

To illustrate the link between meta-learning, causality, OOD generalization and fast
transfer learning, consider the example from \citep{bengio2019meta}. We consider two discrete random variables $A$ and $B$, each taking $N$ possible values. We assume that $A$ and $B$ are correlated, without any hidden confounder. The goal is to determine whether the underlying causal graph is $A \rightarrow B$ ($A$ \emph{causes} $B$), or $B \rightarrow A$. Note that this underlying causal graph cannot be identified from observational data from a single (training) distribution $p$ only, since both graphs are Markov equivalent for $p$, i.e. consistent with observational data of any size. In order to disambiguate between these two hypotheses, \citep{bengio2019meta} use samples from some transfer distribution $\tilde{p}$ in addition to our original samples from the training distribution $p$.

Without loss of generality, they  fix the true causal graph to be $A \rightarrow B$, which is unknown to the learner. Moreover, to make the case stronger, they consider a setting called \emph{covariate shift}, where they assume that the change (again, whose nature is unknown to the learner) between the training and transfer distributions occurs after an intervention on the cause $A$. In other words, the marginal of $A$ changes, while the conditional $p(B\mid A)$ does not, i.e. $p(B\mid A) = \tilde{p}(B\mid A)$. Changes on the cause will be most informative, since they will have direct effects on $B$. ~\citep{bengio2019meta} find experimentally that this is sufficient to identify the causal graph, while~\citep{priol2020analysis} justify this with theoretical arguments in the case where the intervention is on the cause.

In order to demonstrate the advantage of choosing the causal model $A \rightarrow B$ over the anti-causal $B \rightarrow A$, ~\citep{bengio2019meta}  compare how fast the two models can adapt to samples from the transfer distribution $\tilde{p}$. They quantify the speed of adaptation as the log-likelihood after multiple steps of fine-tuning via (stochastic) gradient ascent on the example wise log-likelihood, starting with both models trained on a large amount of data from the training distribution. They show via simulations that the model corresponding to the underlying causal structure adapts faster. Moreover, the difference between the quality of the predictions made by the causal and anti-causal models as they see more post-intervention examples is more significant when adapting on a small amount of data, of the order of 10 to 30 samples from the transfer distribution. Indeed, asymptotically, both models recover from the intervention perfectly and are not distinguishable. This is interesting because it shows that generalization from few examples (after a change in distribution) actually contains more signal about the causal structure than generalization from a lot of examples (whereas in machine learning we tend to think that more data is always better). ~\citep{bengio2019meta} make use of this property (the difference in performance between the two models) as a noisy signal to infer the direction of causality, which here is equivalent to choosing how to modularize the joint distribution. 
The connection to meta-learning is that in the inner loop
of meta-learning we adapt to changes in the distribution, whereas
in the outer loop we slowly converge towards a good model
of the causal structure (which describes what is shared across environments and interventions). Here the meta-parameters capture
the belief about the causal graph structure and the default
(unperturbed) conditional dependencies, while the inner loop parameters
are those which capture the change in the graph due to the intervention.

\citep{ke2019learning} further expanded this idea to deal with more than two variables.  To model causal relations and out-of-distribution generalization one can  
view real-world distributions as arising from the composition of causal mechanisms. Any change in distribution (e.g., when
moving from one setting/domain to a related one) is attributed to changes in as few as possible (but at least one) of those mechanisms~\citep{RIMs, bengio2019meta, ke2019learning}. A do-intervention or hard intervention would set the value of a variable to some value irrespective of the causal parents of that variable, thus disconnecting that node from its parents in the causal graph. By inferring this graph surgery, an intelligent agent should be able to recognize and make sense of these sparse changes and quickly adapt their pre-existing knowledge to this new domain. A current  hypothesis is that a causal graphical model defined on the
appropriate causal variables would be more efficiently learned than one defined on the wrong
representation. Preliminary work based on meta-learning \citep{ke2019learning, dasgupta2019causal, bengio2019meta} suggests that, parameterizing the correct variables and causal structures, the parameters of the graphical model capturing
the (joint) observational distribution can be adapted faster to changes in distribution due to interventions. This comes as a consequence of the fact that fewer parameters need to be adapted to account for the intervention~\citep{priol2020analysis}. In this sense, learning causal representations may bring immediate benefits to machine learning models in terms of reduced
sample complexity.

\vspace{-4mm}
\subsection{Actions and affordances as part of the causal model}

Understanding causes and effects is a crucial component of the human cognitive experience. 
Humans are agents and their actions change the world (sometimes only in little ways), and those
actions can inform them about the causal structure in the world.
Understanding that causal structure is important in order to plan further actions
in order to achieve desired consequences, or to attribute credit to one's or others' actions, i.e.,
to understand and cope with changes in distribution occuring in the world.
However, in realistic settings such as those experienced by a child or a robot, the agent
typically does not have full knowledge of what abstract action was performed and needs to
perform inference over that. The agent would thus have
a causal model of latent causal variables (how they influence
each other and relate to each other), an intervention model
relating low-level actions with interventions (or intentions
to change specific high-level variables), as well as an
observation model (relating high-level causal variables and
sensory observations). In addition to these models, it would have
inference machinery associated with them, including a high-level
policy generating goals (i.e. intentions to intervene in a
particular way).

A human-centric version of this viewpoint is the psychological theory of affordances~\citep{gibson1977theory, cisek2007cortical, pezzulo2016navigating} that can be linked to predictive state representations in reinforcement learning: what can we do with an object? What are the consequences of these actions? Learning affordances as representations of how agents can cause changes in their environment by controlling objects and influencing other agents is more powerful than learning a data distribution. It would not only allow us to predict the consequences of actions we may not have observed at all, but it also allows us to envision which potentialities would result from a different mix of interacting objects and agents. This line of thinking is directly related to the work in machine learning and reinforcement learning on controllability of aspects of the environment~\citep{DBLP:journals/corr/BengioTPPB17, thomas2017independently}. A clue about a good way to define causal variables is precisely that there exist actions or skills to control one causal variable while not directly influencing most others (i.e., except as an effect of the causal variable which is being controlled). A learner thus needs to discover an intervention model (what actions give rise to what interventions), but the locality of interventions in the causal graph can also help the learner figure out a good representation space for causal variables.

\vspace{-4mm}
\section{Conclusions}

To be able to handle dynamic, changing conditions, we want to move from deep statistical models  which are able to perform system 1 tasks to deep structural models also able to perform system 2 tasks
by taking advantage of the computational workhorse of system 1 abilities.  
Today's deep networks may benefit from additional structure and inductive biases to do substantially better
on system 2 tasks, natural language understanding, out-of-distribution systematic generalization
and efficient transfer learning.
We have tried to clarify what some of these inductive biases may be, but much work needs to be
done to improve that understanding and find appropriate ways to incorporate these priors in
neural architectures and training frameworks. We have motivated these inductive biases in terms of expected (and observed in recent work) gains in terms of out-of-distribution generalization and fast adaptation in transfer settings rather than the standard test set from the same distribution as the training set. The general insight here is that the proposed inductive biases should help organize knowledge into the stable reusable parts that are likely to be useful in new settings and tasks (such as causal mechanisms), separating them from the more volatile pieces of information (the values of variables) that can be changed by agents (through causal intervention) or those that are affected by these changes and may vary across environments or tasks.
Some of the more salient inductive biases we propose
deserve especially more attention in deep learning research include (a) the fairly direct connection between high-level variables and natural language or more generally how humans communicate knowledge among them, i.e., we can verbalize our thoughts to a large extent and this can provide rich insights about underlying inductive biases such as these: (b) the modular decomposition of knowledge into independent reusable pieces that can be composed on the fly to address new contexts, (c) the causal interpretation of actions by agents and of changes in distribution, with agents generally intending to affect a single or very few (generally latent) variables, and (d) the sparsity of dependencies between high-level variables (and thus the small number of variables that are linked by causal mechanisms imagined by humans to explain their environment). Finally, we would also like to mention that inductive biases are not the only way to bridge the gap to high-level human cognition: we may gain by improving our optimization algorithms, by scaling up neural networks \citep{sutton2019bitter} and by moving to other frameworks that better capture uncertainty about the world  (e.g., by learning a Bayesian posterior over neural network models as compared to learning a point estimates). It would also be intriguing to think of ways to combine all these different elements together.

\section{Acknowledgements}
The authors are  grateful to Alex Lamb, Rosemary Nan Ke and Olexa Bilaniuk for leading many projects some of which are also discussed here. The authors are grateful to Mike Mozer and Bernhard Schölkopf for many brainstorming discussions.  The authors would also like to thank Stefan Bauer, Aniket Didolkar,  Nasim Rahaman, Kanika Madan, Philippe Beaudoin, Charles Blundell, Dianbo Liu, Moksh Jain for useful feedback. The authors would also like to acknowledge comments by the reviewers which helped in improving the manuscript.

%\begin{thebibliography}{9}
%\bibliographystyle{RS}

%\bibliography{sample}

\bibliography{ref}

%\bibliography{ref}

\end{document}